\documentclass{llncs}

\usepackage{array}
\usepackage{moreverb}
\usepackage{tabularx}
\usepackage{graphicx}
\usepackage{floatflt}
\usepackage{amsmath}
\usepackage{hyperref}

\graphicspath{{.}}

\begin{document}

\mainmatter

\title{Genetic Programming, Validation Sets, and Parsimony Pressure}
\titlerunning{Genetic Programming, Validation Sets, and Parsimony Pressure}

\author{Christian Gagn\'e\inst{1,2,3} \and Marc Schoenauer\inst{1} \and\\
Marc Parizeau\inst{2} \and Marco Tomassini\inst{3}}
\authorrunning{Christian Gagn\'e et al.}
\tocauthor{Christian Gagn\'e (INRIA), Marc Schoenauer (INRIA), Marc Parizeau (Universit\'e Laval), Marco Tomassini (Universit\'e de Lausanne)}
\institute{
\'Equipe TAO -- INRIA Futurs,\\
LRI Bat. 490, Universit\'e Paris Sud, 91405 Orsay CEDEX, France.\\
\email{\{christian.gagne,marc.schoenauer\}@lri.fr}\\[1.2em]
\and
Laboratoire de Vision et Syst\`emes Num\'eriques (LVSN),\\
D\'epartement de G\'enie \'Electrique et de G\'enie Informatique,\\
Universit\'e Laval, Qu\'ebec (QC), G1K 7P4, Canada.\\
\email{parizeau@gel.ulaval.ca}\\[1.2em]
\and
Information Systems Institute,\\
Universit\'e de Lausanne, CH-1015 Dorigny, Switzerland.\\
\email{marco.tomassini@unil.ch}
}

\maketitle

\begin{abstract}
Fitness functions based on test cases are very common in Genetic Programming (GP). This process can be assimilated to a learning task, with the inference of models from a limited number of samples. This paper is an investigation on two methods to improve generalization in GP-based learning: 1) the selection of the best-of-run individuals using a three data sets methodology, and 2) the application of parsimony pressure in order to reduce the complexity of the solutions. Results using GP in a binary classification setup show that while the accuracy on the test sets is preserved, with less variances compared to baseline results, the mean tree size obtained with the tested methods is significantly reduced. 
\end{abstract}

This paper is an experimental study of methodologies for Evolutionary Computations (EC) inspired by common practices in the Machine Learning (ML) and Pattern Recognition (PR) communities. More specifically, using Genetic Programming (GP) for supervised learning, we aim at evaluating both the effect of using a {\em three data sets methodology} (training, validation, and test sets) and the effect of minimizing the classifiers complexity. Our experiments show that these approaches preserve the performances of GP, while significantly reducing the size of the best-of-run solutions, which is in accordance with Occam's Razor principle.

The structure of the paper goes as follow. Section \ref{sec:Introduction} starts with a high-level description of the tested approaches and their justifications. A presentation of relevant work follows in Section \ref{sec:RelatedWork}. Thereafter, the methodology used in the experiments is detailed in Section \ref{sec:Methodology}. Finally, Section \ref{sec:Results} presents the experimental results obtained on six binary classification data sets, and Section \ref{sec:Conclusion} concludes the paper.

\section{Introduction}
\label{sec:Introduction}

GP is particularly suited for problems that can be assimilated to learning tasks, with the minimization of the error between the obtained and desired outputs for a limited number of test cases -- the training data, using a ML terminology. Indeed, the classical GP examples of symbolic regression, boolean multiplexer and artificial ant \cite{Koza1992} are only simple instances of well-known learning problems (i.e. respectively regression, binary classification and reinforcement learning). In the early years of GP, these problems were tackled using a single data set, reporting results on the same data set that was used to evaluate the fitnesses during the evolution. This was justifiable by the fact that these are toy problems used only to illustrate the potential of GP. In the ML community, it is recognized that such methodology is flawed, given that the learning algorithm can overfit the data used during the training and perform poorly on unseen data of the same application domain \cite{Duda2001,Mitchell1997}. Hence, it is important to report results on a set of data that was not used during the learning stage. This is what we call in this paper a {\em two data sets methodology}, with a training set used by the learning algorithm and a test set used to report the performance of the algorithm on unseen data, which is a good indicator of the algorithm's generalization (or robustness) capability. Even though this methodology has been widely accepted and applied in the ML and PR communities for a long time, the EC community still lags behind by publishing papers that are reporting results on data sets that were used during the evolution (training) phase. This methodological problem has already been spotted (see \cite{Eiben2002}) and should be less and less common in the future.

The two data sets methodology prevents reporting flawed results of learning algorithms that overfit the training set. But this does not prevent by itself overfitting the training set. A common approach is to add a third data set -- the validation set -- which helps the learning algorithm to measure its generalization capability. This validation set is useful to interrupt the learning algorithm when overfitting occurs and/or select a configuration of the learning machine that maximizes the generalization performances. This third data set is commonly used to train classifiers such as back-propagation neural networks and can be easily applied to EC-based learning. But this approach has an important drawback: it removes a significant amount of data from the training set, which can be harmful to the learning process. Indeed, the richer the training set, the more representative it can be of the real data distribution, and the more the learning algorithm can be expected to converge toward robust solutions. In the light of these considerations, an objective of this paper is to investigate the effect of a validation set to select the best-of-run individuals for a GP-based learning application.

Another concern of the ML and PR communities is to develop learning algorithms that generate simple solutions. An argument behind this is the Occam's Razor principle, which states that between solutions of comparable quality, the simplest solutions must be preferred. Another argument is the minimum description length principle \cite{Rissanen78}, which states that the ``best'' model is the one that minimizes the amount of information needed to encode the model and the data given the model. Preference for simpler solutions and overfitting avoidance are closely related: it is more likely that a complex solution incorporates specific information from the training set, thus overfitting the training set, compared to a simpler solution. But, as mentioned in \cite{Domingos1999}, this argumentation should be taken with care as too much emphasis on minimizing complexity can prevent the discovery of more complex yet more accurate solutions.

There is a strong link between the minimization of complexity in GP-based learning and the control of code bloat \cite{Koza1992,Banzhaf2002}, that is an exaggerated growth of program size in the course of GP runs. Even though complexity and code bloat are not exactly the same phenomenon, as some kind of bloat is generated by neutral pieces of code that have no effect on the actual complexity of the solutions, most of the mechanisms proposed to control it \cite{Langdon2000,Ekart2001,Luke2002,Silva2003} can also be used to minimize the complexity of solutions obtained by GP-based learning.

This paper is a study of GP viewed as a learning algorithm. More specifically, we investigate two techniques to increase the generalization performance and decrease the complexity of the models: 1) use of a validation set to select best-of-run individuals that generalize well, and 2) use of lexicographic parsimony pressure \cite{Luke2002} to reduce the complexity of the generated models. These techniques are tested using a GP encoding for binary classification problems, with vectors taken from the learning sets as terminals, and mathematical operations to manipulate these vectors as branches. This approach is tested on six different data sets from the UCI ML repository \cite{Newman1998}. Even if the proposed techniques are tested in a specific context, we argue that they can be extended to the frequent situations where GP is used as a learning algorithm.

\section{Related Work}
\label{sec:RelatedWork}

Some GP learning applications \cite{Sherrah1997,Brameier2001,Yu2004} have made use of a three data sets methodology, but without making a thorough analysis of its effects. Panait and Luke \cite{Panait2003} conducted some experiments on different approaches to increase the robustness of the solutions generated by GP, using a three data sets methodology to evaluate the efficiency of each approach. Rowland \cite{Rowland2003} and Kushchu \cite{Kushchu2002} conducted studies on generalization in EC and GP. Both of their argumentations converge toward the testing of solutions in previously unseen situations for improving robustness.

Because of the bloat phenomenon, typical in GP, parsimony pressure has been more widely studied \cite{Ekart2001,Nordin1995,Soule1998,Gustafson2004}. In particular, several papers \cite{Iba1994,Zhang1995,Rosca1996} have produced interesting results around the idea of using a parsimony pressure to increase the generalization capability of GP-evolved solutions. However, a counter-argumentation is given in \cite{Cavaretta1999}, where solutions biased toward low complexity have, in some circumstances, increased generalization error. This is in accordance with the argumentation given in \cite{Domingos1999}, which states that less complex solutions are not always more robust.

\section{Methodology}
\label{sec:Methodology}

The experiments conducted in this work are based on a GP-setup specialized for binary classification problems. The data processed by the primitives are vectors of two possible sizes, either of size one (a scalar value), or of size $n$, the feature set size. Table \ref{tab:GPPrimitives} presents the set of primitives used to build the programs.
\begin{table}[t]
\caption[GP primitives used to build the classifiers.]{GP primitives used to build the classifiers.}
\label{tab:GPPrimitives}
\begin{center}
\begin{tabularx}{\linewidth}{c|c|X}
Name & \# args. & Description \\\hline
ADD  & $2$      & Addition, $\mathrm{f_{ADD}}(x_1,x_2)=x_1+x_2$.\\ 
SUB  & $2$      & Subtraction, $\mathrm{f_{SUB}}(x_1,x_2)=x_1-x_2$.\\
MUL  & $2$      & Multiplication, $\mathrm{f_{MUL}}(x_1,x_2)=x_1 x_2$.\\
DIV  & $2$      & Protected division, $\mathrm{f_{DIV}}(x_1,x_2)=\left\{\begin{array}{c@{\qquad}c}1 & |x_2|<0.001\\ x_1/x_2 & \mbox{otherwise}\end{array}\right.$.\\
MXF  & $2$      & Maximum value, $\mathrm{f_{MXF}}(x_1,x_2)=\max(x_1,x_2)$.\\
MNF  & $2$      & Minimum value, $\mathrm{f_{MNF}}(x_1,x_2)=\min(x_1,x_2)$.\\
ABS  & $1$      & Absolute value, $\mathrm{f_{ABS}}(x)=|x|$.\\
SLN  & $1$      & Saturated symmetric linear function, $\mathrm{f_{SLN}}(x)=\left\{\begin{array}{c@{\qquad}c}1 & x>1\\ -1 & x<-1\\ x & \mbox{otherwise}\end{array}\right.$.\\
SUM  & $1$      & Sum of vector's components, $\mathrm{f_{SUM}}(\mathbf{x})=\sum_i{x_{i}}$.\\
MEA  & $1$      & Mean of vector's components, $\mathrm{f_{MEA}}(\mathbf{x})=\frac{{\sum_i}{x_{i}}}{{\mathrm{card}(\mathbf{x})}}$.\\
MXV  & $1$      & Maximum of vector's components, $\mathrm{f_{MXV}}(\mathbf{x})=\max_i{x_{i}}$.\\
MIV  & $1$      & Minimum of vector's components, $\mathrm{f_{MIV}}(\mathbf{x})=\min_i{x_{i}}$.\\
L2   & $1$      & $L_2$ norm of the vector, $\mathrm{f_{L2}}(\mathbf{x})=\sqrt{\sum_i{x_{i}^2}}$.\\
E    & $0$      & Ephemeral random vector, generated by copying the value of a randomly selected training set data.\\
X    & $0$      & Vector with the value of the data to classify.\\
\end{tabularx}
\end{center}
\end{table}

Three main families of primitives were used: the {\em mathematical function primitives} (ADD, SUB, MUL, DIV, MXF, MNF, ABS, and SLN), the {\em vector-to-scalar primitives} (SUM, MEA, MXV, MIV, and L2), and the {\em vectorial terminals} (E and X). The mathematical function primitives with two arguments (ADD, SUB, MUL, DIV, MXF, and MIF) are defined to deal with arguments of different sizes by applying the function to each component of the $n$-sized arguments, when necessary repeatedly using the value of the scalar arguments. More formally, if $f(x_1,x_2)$ denotes the function associated to the primitive presented in Table \ref{tab:GPPrimitives}, the output of these primitives is:
\begin{itemize}
\item A scalar $[\mathrm{f}(x_1(1),x_2(1))]$, if both arguments are scalars;
\item A size $n$ vector $[\mathrm{f}(x_1(1),x_2(1))~~\mathrm{f}(x_1(1),x_2(2))~\ldots~\mathrm{f}(x_1(1),x_2(n))]^T$, if the first argument is a scalar and the second a vector;
\item A size $n$ vector $[\mathrm{f}(x_1(1),x_2(1))~~\mathrm{f}(x_1(2),x_2(1))~\ldots~\mathrm{f}(x_1(n),x_2(1))]^T$, if the first argument is a vector and the second a scalar;
\item A size $n$ vector $[\mathrm{f}(x_1(1),x_2(1))~~\mathrm{f}(x_1(2),x_2(2))~\ldots~\mathrm{f}(x_1(n),x_2(n))]^T$, if both arguments are vectors.
\end{itemize}
On the other hand, the vector-to-scalar primitives are defined to convert a vector argument of size $n$ into a scalar output. When the argument is a scalar, it is returned as output value as is, without modification, except for the L2 primitive which returns the absolute value of the input scalar. Finally, the vectorial terminals are always vectors of size $n$, with either randomly selected data of the training set (terminal E), used as constants, or the value of the data to classify (terminal X), used as the variable of the problem. 

The data evaluated is classified according to the output of the GP tree, that is assigned to the first class for an output value positive or zero, otherwise assigned to the second class. If necessary, the output of the GP program is converted into a scalar beforehand, by a summation of each vector's components, as does the primitive SUM.

In order to test the effect of using a validation set and applying some parsimony pressure, GP will be tested on common binary classification data sets taken from the {\em Machine Learning Repository} at UCI \cite{Newman1998}. The selected data set are presented in Table \ref{tab:UCIDataSetDescription}.
\begin{table}[t]
\caption{Description of UCI data sets used for the experimentations.}
\label{tab:UCIDataSetDescription}
\begin{center}
\begin{tabularx}{\linewidth}{c|c|c|X}
Data  &        & \# of    &\\
set   & Size   & features & Application domain\\\hline
bcw   & $699$  & $9$      & Wisconcin's breast cancer, $65.5\:\%$ benign and $34.5\:\%$ malignant.\\
cmc   & $1473$ & $9$      & Contraceptive method choice, $42.7\:\%$ not using contraception and $57.3\:\%$ using contraception.\\
ger   & $1000$ & $24$     & German credit approval, $70\:\%$ approved and $30\:\%$ not approved.\\
ion   & $351$  & $34$     & Ionosphere radar signal, $35.9\:\%$ without structure detected and $64.9\:\%$ with a structure detected.\\
pid   & $768$  & $8$      & Pima Indians diabetes, $65.1\:\%$ tested negative and $34.9\:\%$ tested positive for diabetes.\\
spa   & $4601$ & $57$     & Spam e-mail, $60.6\:\%$ non-junk e-mail and $39.4\:\%$ junk e-mail.\\
\end{tabularx}
\end{center}
\end{table}
The selection of these data sets was guided by the following main criteria: 1) select appropriate sets for binary classification, 2) select appropriate sets for $10$-folds cross-validation (see below), that is data sets without predefined separated training and testing sets, and 3) select sets of relatively large size or high dimensionality. The first two criteria were chosen in order to fit into our general methodology, to avoid special data manipulations, while the last criterion was postulated in an attempt to select not too easy data sets, that should help to generate discriminant results.

Before the experiments, each data set was randomly divided into $10$ folds of equal size, taking care to balance the number of data of each class between the folds. A $10$-folds cross-validation \cite{Duda2001} has been conducted using the data in $9$ folds as the training set for an evolution, reporting the test set error rate on the remaining fold. For each tested configuration, the process is repeated $10$ times for each fold, for a total of $100$ evolutions per configuration. The reported results consist in the means for these $100$ evolutions.

Our experimentations are conducted on four different configurations:
\begin{enumerate}
\item {\bf Baseline}: The fitness measure consists in minimizing the error rate on the complete training set. The best-of-run individual is simply the individual of the evolution with the lowest error rate on the training set, with the smallest individual selected in cases of ties.
\item {\bf With validation}: For each evolution, the training set is randomly divided into two data sets: the fitness evaluation data set, with $67\%$ of the training data, and the validation set, with the remaining $33\%$. The class distribution of the data is well-balanced between the sets. The fitness measure consists in minimizing the error rate on the fitness evaluation set. At each generation, a two-objective sort is conducted in order to extract a set of non-dominated individuals (the Pareto front), with regards to the lowest fitness evaluation set error rate and the smallest individuals. These non-dominated individuals are then evaluated on the validation set, with the best-of-run individual selected as the one of these with the smallest error rate on the validation set, ties being solved by choosing the smallest individual.
\item {\bf With parsimony pressure}: A lexicographic parsimony pressure \cite{Luke2002} is applied to the evolution by minimizing the error rate on the complete training set, using the individual size as second point of comparison in cases of identical error rates. As with the baseline configuration, the best-of-run individual is the individual of the evolution with the lowest error rate on the training set, with the smallest individual selected in cases of ties (strict equality).
\item {\bf With validation and parsimony pressure}: A mix of the two previous configurations, by separating the training set into two sets, the fitness evaluation set ($67\%$ of the data) and the validation set ($33\%$ of the data), and making use of the lexicographic parsimony pressure. The fitness evaluation set is used to compute the error rate that guides the evolution while the validation set is used only to select the best-of-run individual. The selection of this best-of-run individual is identical to the {\em with validation} configuration, by extracting a Pareto front of the non-dominated individuals of the current generation (fitness evaluation set error rates vs individual sizes). At each generation, all these non-dominated individuals are tested on the validation set. The best-of-run individual is selected as the solution that minimizes the validation error rate, breaking ties by preferring the smallest individuals. 
\end{enumerate}

Thus, for the second and fourth settings, the Pareto front is extracted at each generation for testing against the validation set. This is motivated by two main reasons: 1) it is important to reduce the number of solutions tested against the validation set, in order not to select best-of-run solutions that are just ``by chance'' performing well on the validation set, and 2) it is desirable to test on the validation set a range of solutions with different accuracy/size trade-offs. It should be stressed that tournament selection is used in all evolutions, with lexicographic ranking for the third and fourth configurations. Strictly speaking, this is not a Pareto domination-based multi-objective selection algorithm.

Table \ref{tab:GPParameters} presents the GP parameters used during the experiments.
\begin{table}[t]
\caption{Tableau of the GP evolutions.}
\label{tab:GPParameters}
\begin{center}
\begin{tabularx}{\linewidth}{c|X}
Parameter              & Description and parameter values \\\hline
Terminals and branches & See Table \ref{tab:GPPrimitives}.\\
Population size        & One panmictic population of $1000$ individuals.\\
Stop criterion         & Evolution ends after $100$ generations.\\
Replacement strategy   & Genetic operations applied following generational scheme.\\
Selection              & Tournaments selection with $2$ participants (relative ranking).\\
Fitness measure        & {\bf Without parsimony pressure:} minimize the error rate.\\
                       & {\bf With parsimony pressure:} minimize the error rate and, in case of ties, select the smallest individuals (lexicographic ranking).\\
Crossover              & Classical subtree crossover \cite{Koza1992} (probability $0.7$).\\
Standard mutation      & Replace a subtree with a new randomly generated one (probability $0.05$).\\
Swap mutation          & Exchange a primitive with another of the same arity (probability $0.05$).\\
Shrink mutation        & Replace a branch with one of its children and remove the branch mutated and the other children's subtrees (if any) (probability $0.05$).\\
Ephemerals mutation    & Randomly select a new ephemeral random vector (probability $0.05$).\\
Reproduction           & Copy without modification an existing individual (probability $0.1$).\\
Data normalization     & The data of the different sets are scaled in $[-1,1]$ along the different dimensions.\\
\end{tabularx}
\end{center}
\end{table}
No special tweaking of these parameter values was done, which correspond in most cases to the default values of the software tool used. The experimentations have been implemented using the GP facilities of the Open BEAGLE framework \cite{Gagne2002}.

\section{Results}
\label{sec:Results}

Table \ref{tab:DetailedResults} presents the detailed results obtained by testing the four configurations presented in the previous section, using the six data sets of Table \ref{tab:UCIDataSetDescription}.
\begin{table}
\caption[Error rates, tree sizes and efforts for the evolution of GP-based classifiers using the UCI data sets.]{Error rates, tree sizes and effort for the evolution of GP-based classifiers using the UCI data sets. Results in {\it italic} are not statistically different from those of the baseline configuration, according to a $95\%$ confidence two-tailed Student's $t$-test. Results in {\bf bold} are more than $50\%$ smaller than the corresponding baseline results.}
\label{tab:DetailedResults}
\begin{center}
\begin{tabular}{c|c|c|c|c|c|c|c|c|c|c}
           & \multicolumn{2}{c|}{Train set rate} & \multicolumn{2}{c|}{Valid. set rate} & \multicolumn{2}{c|}{Test set rate} & \multicolumn{2}{c|}{Tree size} & \multicolumn{2}{c}{Effort}\\
           & Mean                 & Std.         & Mean                 & Std.          & Mean                 & Std.        & Mean             & Std.        & Mean            & Stdev.\\
Approach   & error                & dev.         & error                & dev.          & error                & dev.        & size             & dev.        & ($\times10^9$)  & ($\times10^9$)\\\hline
\multicolumn{11}{c}{bcw}\\\hline
Baseline   & $1.7\:\%$            & $0.5\:\%$    & --                   & --            & $3.4\:\%$            & $2.3\:\%$   & $83.4$           & $55.2$      & $4.92$          & $1.5$\\
Validation & $2.3\:\%$            & $0.7\:\%$    & $2.3\:\%$            & $0.8\:\%$     & $\mathit{3.3\:\%}$   & $2.3\:\%$   & $\mathbf{34.2}$  & $38.8$      & $4.08$          & $1.2$\\
Parsimony  & $2.1\:\%$            & $0.5\:\%$    & --                   & --            & $\mathit{3.5\:\%}$   & $2.3\:\%$   & $\mathbf{22.0}$  & $18.9$      & $1.10$          & $0.83$\\
Both       & $2.8\:\%$            & $0.7\:\%$    & $2.7\:\%$            & $1.0\:\%$     & $\mathit{3.3\:\%}$   & $2.1\:\%$   & $\mathbf{6.5}$   & $11.2$      & $0.72$          & $0.55$\\\hline
\multicolumn{11}{c}{cmc}\\\hline
Baseline   & $26.3\:\%$           & $2.2\:\%$    & --                   & --            & $31.2\:\%$           & $4.8\:\%$   & $174.8$          & $68.2$      & $11.2$          & $3.5$\\
Validation & $28.6\:\%$           & $3.2\:\%$    & $30.8\:\%$           & $3.0\:\%$     & $\mathit{32.5\:\%}$  & $4.5\:\%$   & $106.4$          & $68.3$      & $8.43$          & $2.7$\\
Parsimony  & $27.0\:\%$           & $2.8\:\%$    & --                   & --            & $\mathit{31.7\:\%}$  & $4.9\:\%$   & $151.6$          & $62.4$      & $10.1$          & $3.9$\\
Both       & $29.3\:\%$           & $3.0\:\%$    & $29.6\:\%$           & $3.0\:\%$     & $\mathit{32.1\:\%}$  & $5.0\:\%$   & $\mathbf{63.7}$  & $39.8$      & $6.17$          & $2.2$\\\hline
\multicolumn{11}{c}{ger}\\\hline
Baseline   & $22.7\:\%$           & $1.6\:\%$    & --                   & --            & $29.3\:\%$           & $3.8\:\%$   & $175.3$          & $77.9$      & $7.43$          & $2.7$\\
Validation & $25.3\:\%$           & $2.6\:\%$    & $27.3\:\%$           & $1.5\:\%$     & $\mathit{29.5\:\%}$  & $3.5\:\%$   & $\mathbf{78.2}$  & $68.8$      & $5.13$          & $1.6$\\
Parsimony  & $\mathit{22.6\:\%}$  & $1.7\:\%$    & --                   & --            & $\mathit{29.1\:\%}$  & $3.8\:\%$   & $141.8$          & $69.2$      & $5.73$          & $2.6$\\
Both       & $25.7\:\%$           & $2.7\:\%$    & $26.7\:\%$           & $1.6\:\%$     & $\mathit{29.6\:\%}$  & $3.2\:\%$   & $\mathbf{54.8}$  & $47.1$      & $3.79$          & $2.0$\\\hline
\multicolumn{11}{c}{ion}\\\hline
Baseline   & $4.1\:\%$            & $1.2\:\%$    & --                   & --            & $10.5\:\%$           & $5.4\:\%$   & $149.4$          & $53.0$      & $2.80$          & $0.76$\\
Validation & $5.9\:\%$            & $3.1\:\%$    & $7.5\:\%$            & $3.5\:\%$     & $\mathit{11.3\:\%}$  & $6.8\:\%$   & $94.2$           & $56.3$      & $2.08$          & $0.55$\\
Parsimony  & $\mathit{4.2\:\%}$   & $1.3\:\%$    & --                   & --            & $\mathit{10.1\:\%}$  & $6.0\:\%$   & $84.4$           & $38.8$      & $1.88$          & $0.59$\\
Both       & $7.7\:\%$            & $2.9\:\%$    & $7.5\:\%$            & $2.8\:\%$     & $\mathit{11.0\:\%}$  & $6.3\:\%$   & $\mathbf{41.6}$  & $28.3$      & $\mathbf{1.10}$ & $0.35$\\\hline
\multicolumn{11}{c}{pid}\\\hline
Baseline   & $19.9\:\%$           & $1.2\:\%$    & --                   & --            & $25.2\:\%$           & $4.5\:\%$   & $149.5$          & $56.8$      & $5.47$          & $1.6$\\
Validation & $22.0\:\%$           & $2.1\:\%$    & $22.9\:\%$           & $2.2\:\%$     & $\mathit{25.2\:\%}$  & $4.5\:\%$   & $\mathbf{60.4}$  & $55.5$      & $4.25$          & $1.3$\\
Parsimony  & $20.1\:\%$           & $1.2\:\%$    & --                   & --            & $\mathit{24.7\:\%}$  & $4.4\:\%$   & $99.6$           & $59.0$      & $3.85$          & $1.2$\\
Both       & $23.5\:\%$           & $2.0\:\%$    & $22.4\:\%$           & $2.0\:\%$     & $\mathit{25.1\:\%}$  & $4.4\:\%$   & $\mathbf{28.0}$  & $25.4$      & $\mathbf{2.45}$ & $0.89$\\\hline
\multicolumn{11}{c}{spa}\\\hline
Baseline   & $12.8\:\%$           & $2.2\:\%$    & --                   & --            & $13.6\:\%$           & $2.7\:\%$   & $166.6$          & $62.4$      & $34.4$          & $9.4$\\
Validation & $\mathit{12.9\:\%}$  & $2.3\:\%$    & $13.7\:\%$           & $2.7\:\%$     & $\mathit{13.9\:\%}$  & $2.6\:\%$   & $148.7$          & $61.7$      & $21.7$          & $6.6$\\
Parsimony  & $\mathit{13.3\:\%}$  & $2.6\:\%$    & --                   & --            & $\mathit{14.2\:\%}$  & $3.2\:\%$   & $141.3$          & $56.4$      & $28.6$          & $10.1$\\
Both       & $\mathit{13.1\:\%}$  & $2.2\:\%$    & $13.5\:\%$           & $2.2\:\%$     & $\mathit{13.9\:\%}$  & $2.5\:\%$   & $109.3$          & $47.0$      & $18.7$          & $6.4$\\\hline
\end{tabular}
\end{center}
\end{table}
The error rates and tree sizes that are reported consist in the mean and standard deviation values of the best-of-run individuals for the $100$ runs ($10$ different runs for each folds). The {\em effort}\footnote{Note that the notion of ``effort'' presented here is different from the one defined by Koza in \cite{Koza1992}.} consists in a measure of the computations done during the evolutions. It is calculated by summing the number of GP primitives evaluated during the runs. More precisely, for configurations without validation, the effort is computed by counting in the number of primitives in each individual times the training set size, for all evaluated individuals during the run. For configurations with validation, the size of the individuals on Pareto front times the validation set size is also taken into account. Italic results in Table \ref{tab:DetailedResults} are not statistically different from the corresponding baseline result; hence all other results are statistically distinct from the baseline.

\begin{figure}[t]
\caption[One-way analysis of variance (ANOVA) box plots of the best-of-run solutions test set error rates.]{One-way analysis of variance (ANOVA) box plots of the best-of-run solutions test set error rates. The center box is bounded by the first and third quartiles of the data distribution, with the median as the central line in the box. The notches surrounding the median show the $95\%$ confidence interval of this median. The whiskers above and below the boxes represent the spread of the data value within $1.5$ times the interquartile range, with the $+$ symbol showing outliers.}
\label{fig:TestRates}
\begin{center}
\begin{tabular}{ccc}
\includegraphics[width=0.33\linewidth]{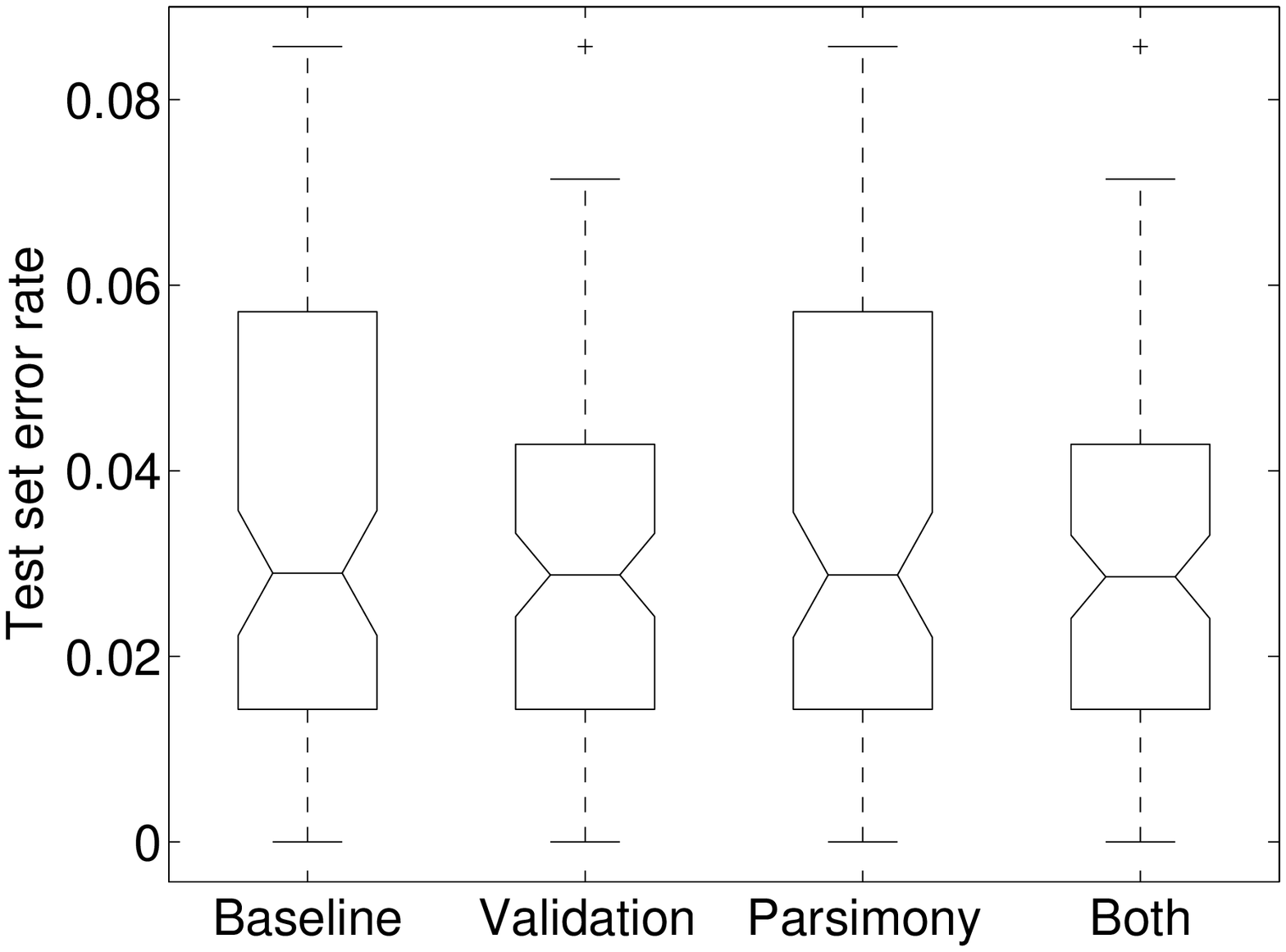} & 
\includegraphics[width=0.33\linewidth]{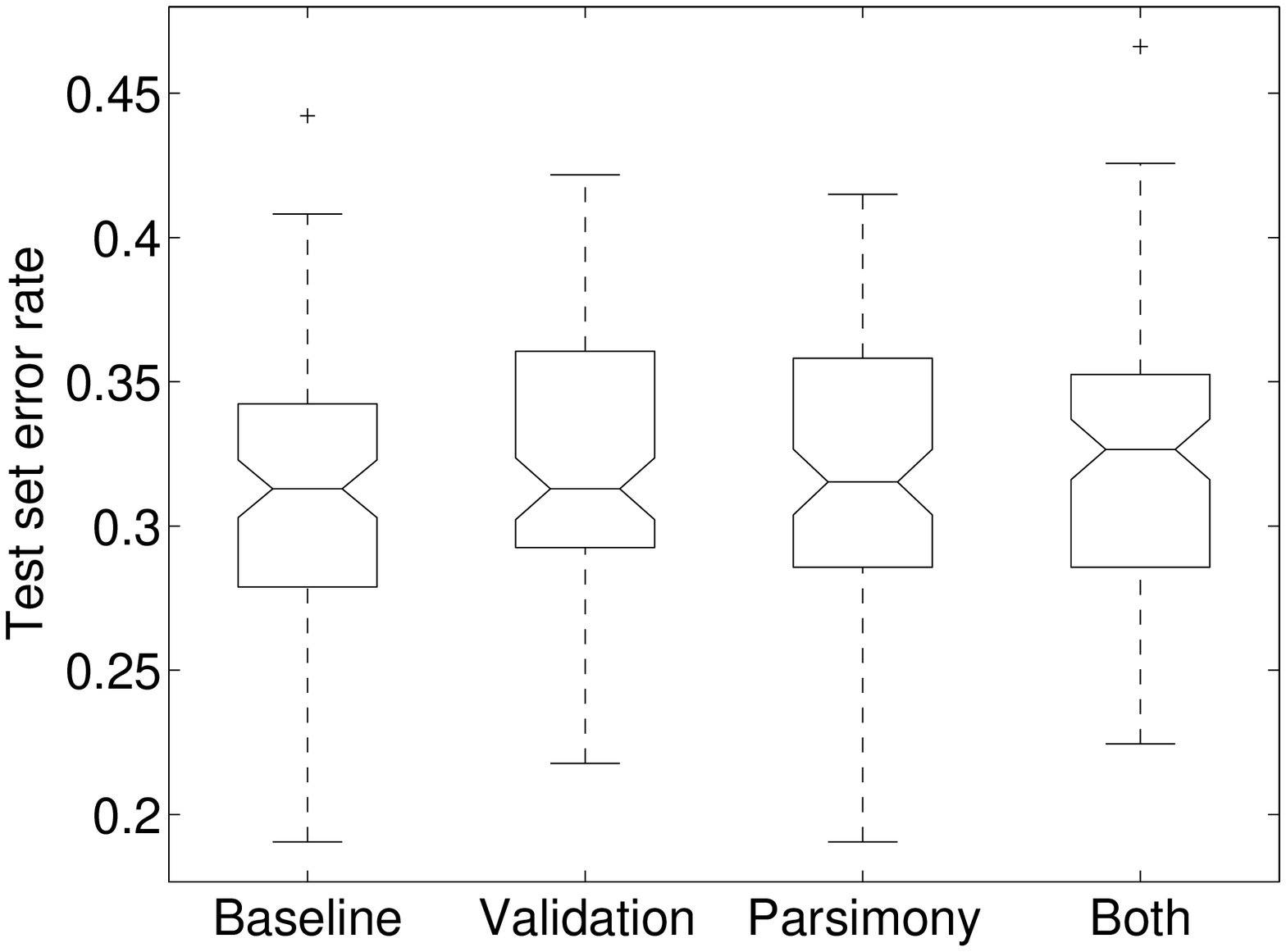} & 
\includegraphics[width=0.33\linewidth]{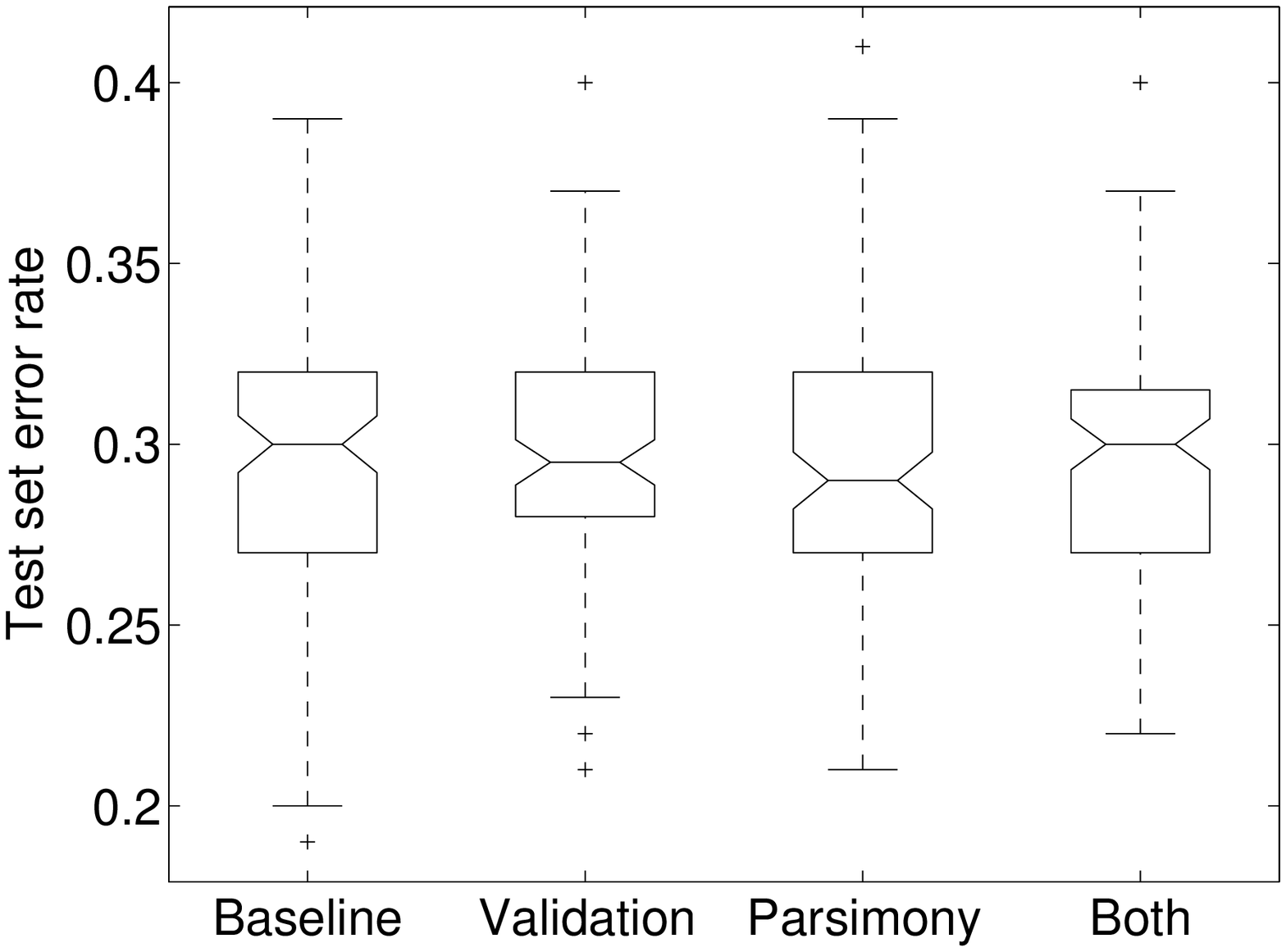}\\
bcw & cmc & ger\\
\includegraphics[width=0.33\linewidth]{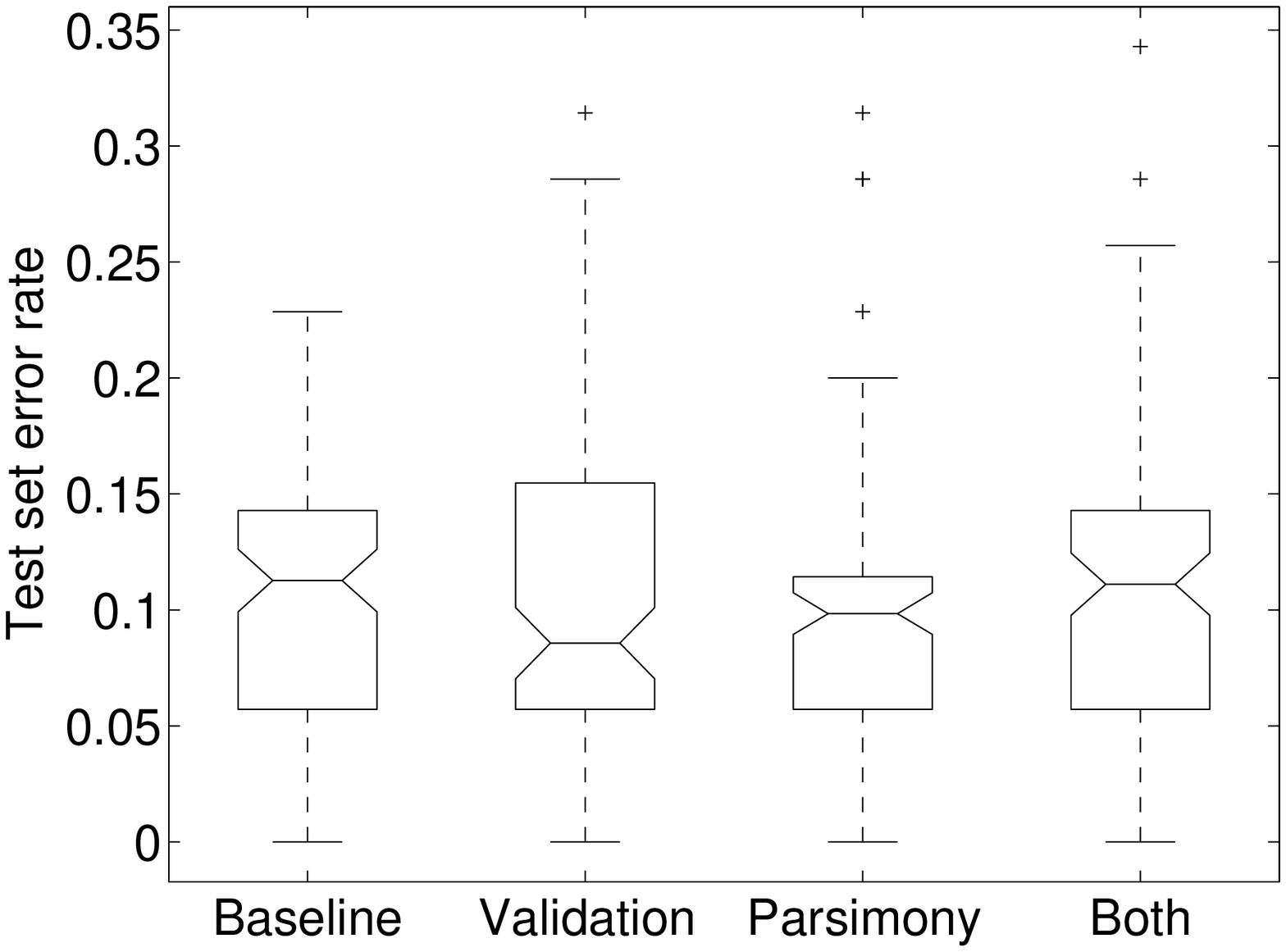} &
\includegraphics[width=0.33\linewidth]{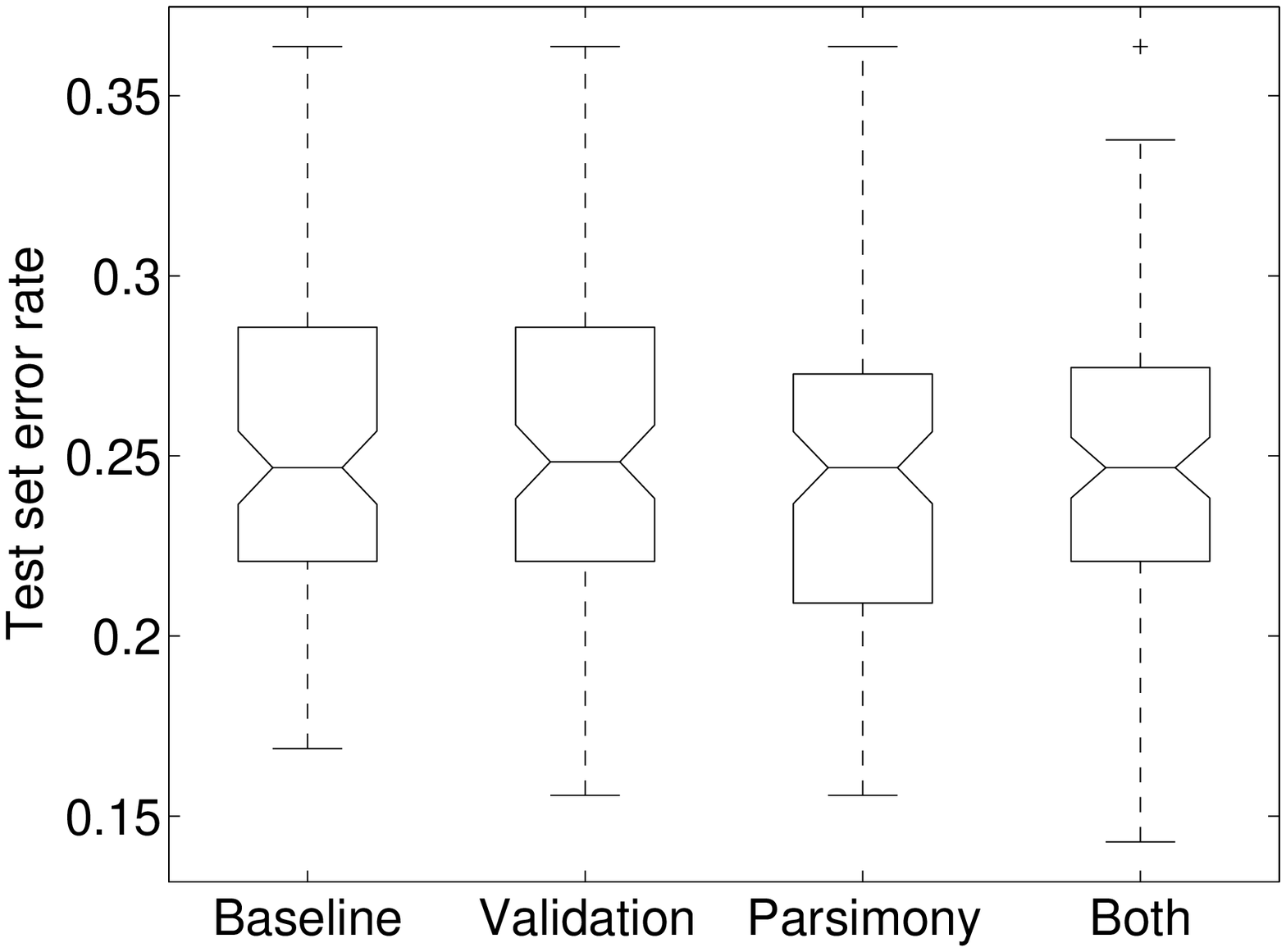} & 
\includegraphics[width=0.33\linewidth]{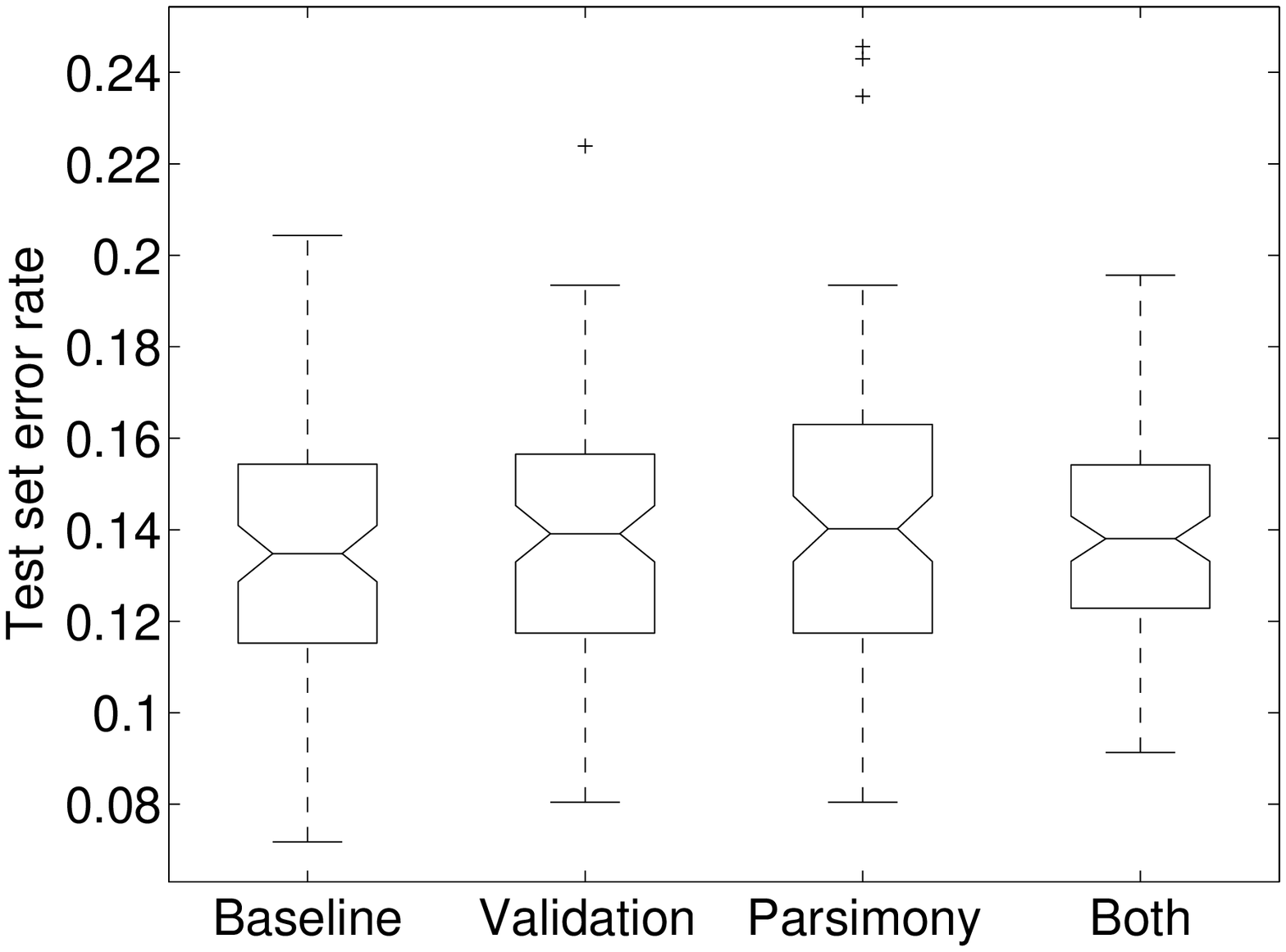}\\
ion & pid & spa\\
\end{tabular}
\end{center}
\end{figure}

Figure \ref{fig:TestRates} presents the box plots that stem from a one-way analysis of variance (ANOVA) conducted on the test set error rates.
Looking at the results, it seems that no approach is clearly superior to the others in term of test set accuracy. But, taking a closer look we can see that the approach using both the validation set and parsimony pressure reduces the variance of the test set error rates (first to third quartile range) for the bcw, ger, pid and spa data sets, having a comparable or slightly worse variance for the two other sets. This is an important result as getting reproducible and stable solutions is often more interesting than finding only infrequently a marginally better individual.

Taking a closer look at the error rates on the different sets in Table \ref{tab:DetailedResults}, important differences can be noted between the train and validation set rates, on one hand, and the test set rates on the other hand. The differences between the train and test rates can be explained by an overfitting of the training data. But, it is surprising to see the importance of the differences between the validation and test rates. This may indicate that, because too many solutions are still tested against the validation set at each generation, the risk of selecting solutions that fit the validation set ``by chance'' is not negligible.

\begin{figure}[t]
\caption{One-way analysis of variance (ANOVA) box plots of the best-of-run solutions tree sizes.}
\label{fig:TreeSizes}
\begin{center}
\begin{tabular}{ccc}
\includegraphics[width=0.33\linewidth]{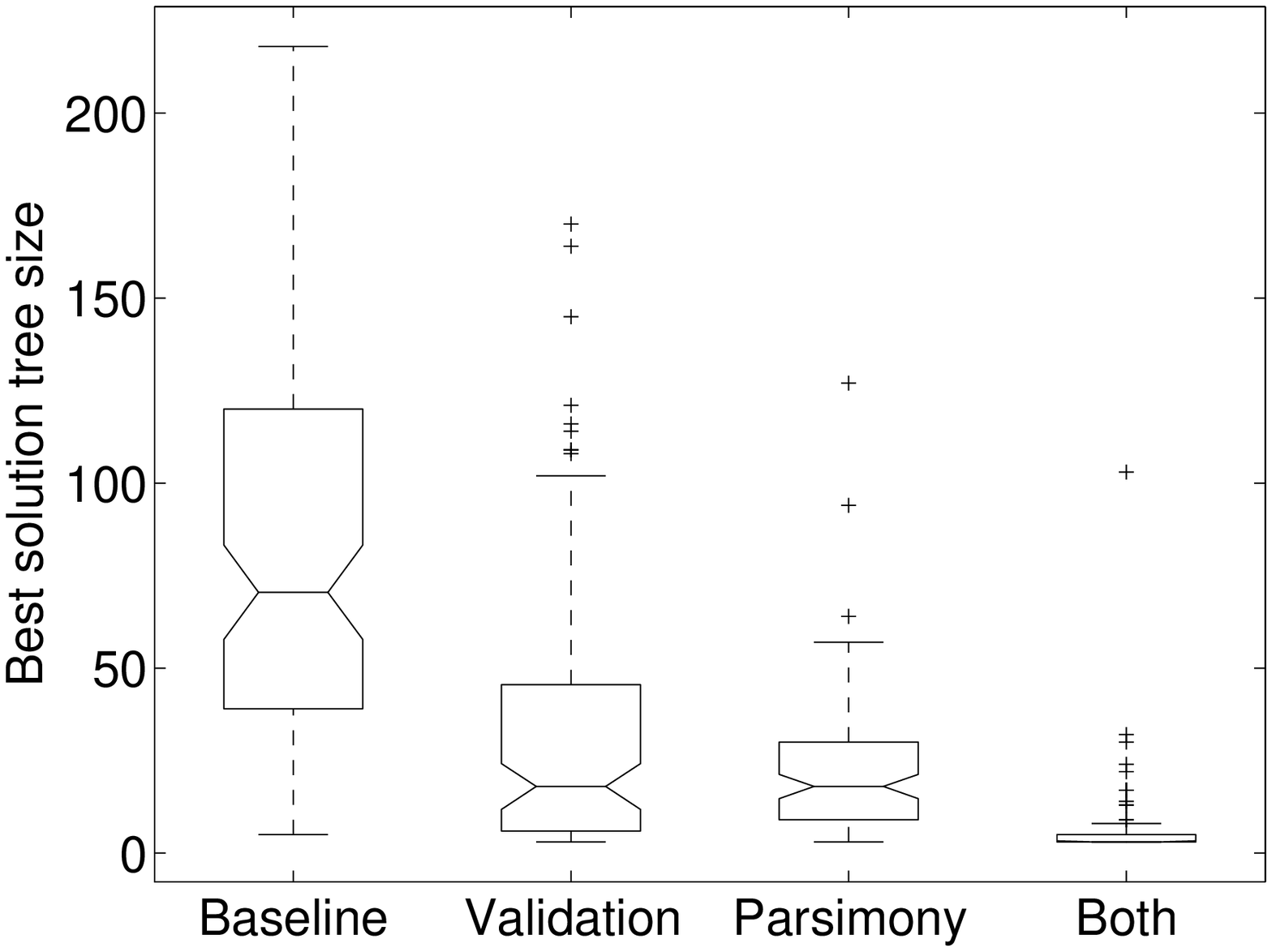} &
\includegraphics[width=0.33\linewidth]{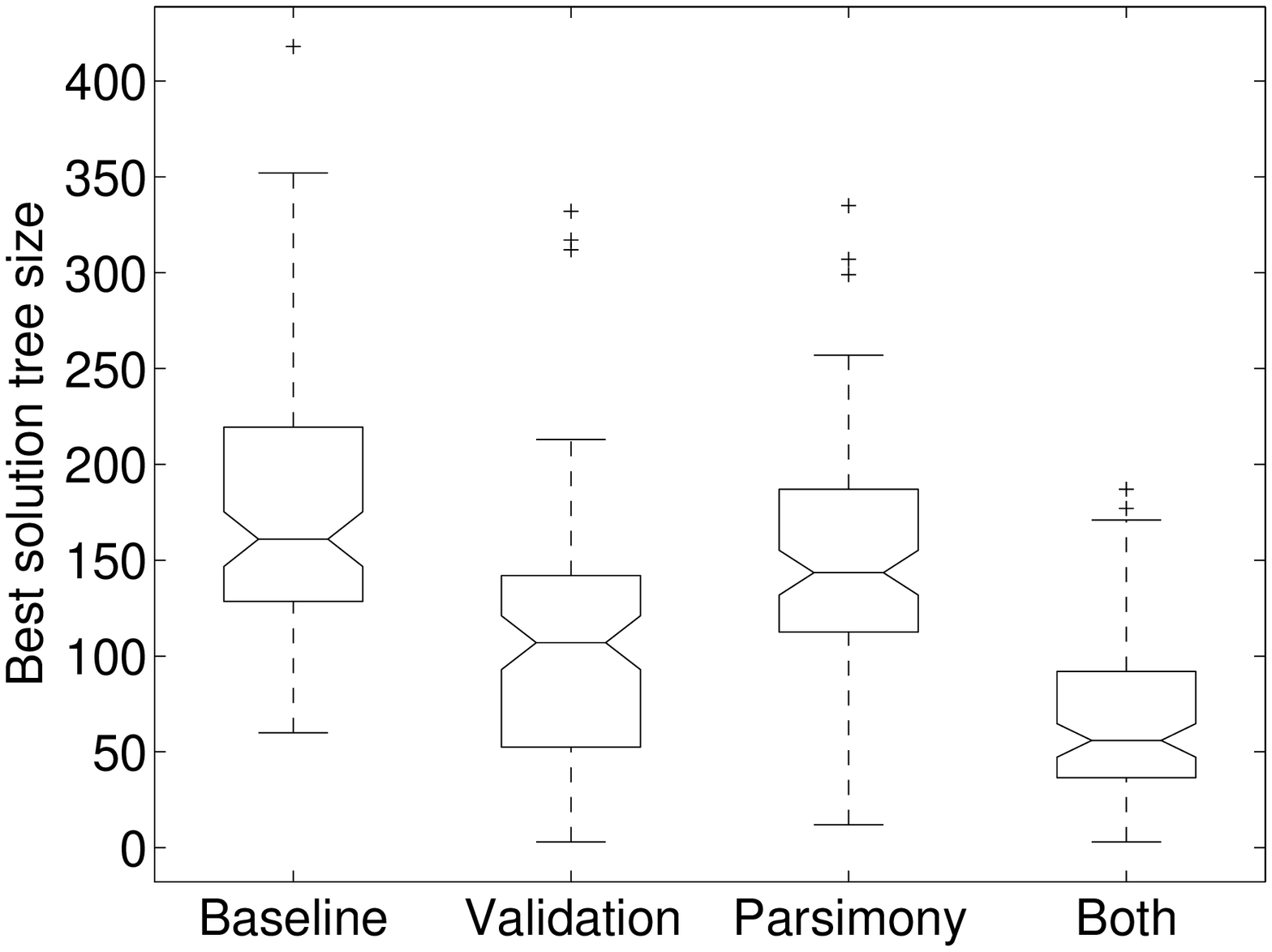} &
\includegraphics[width=0.33\linewidth]{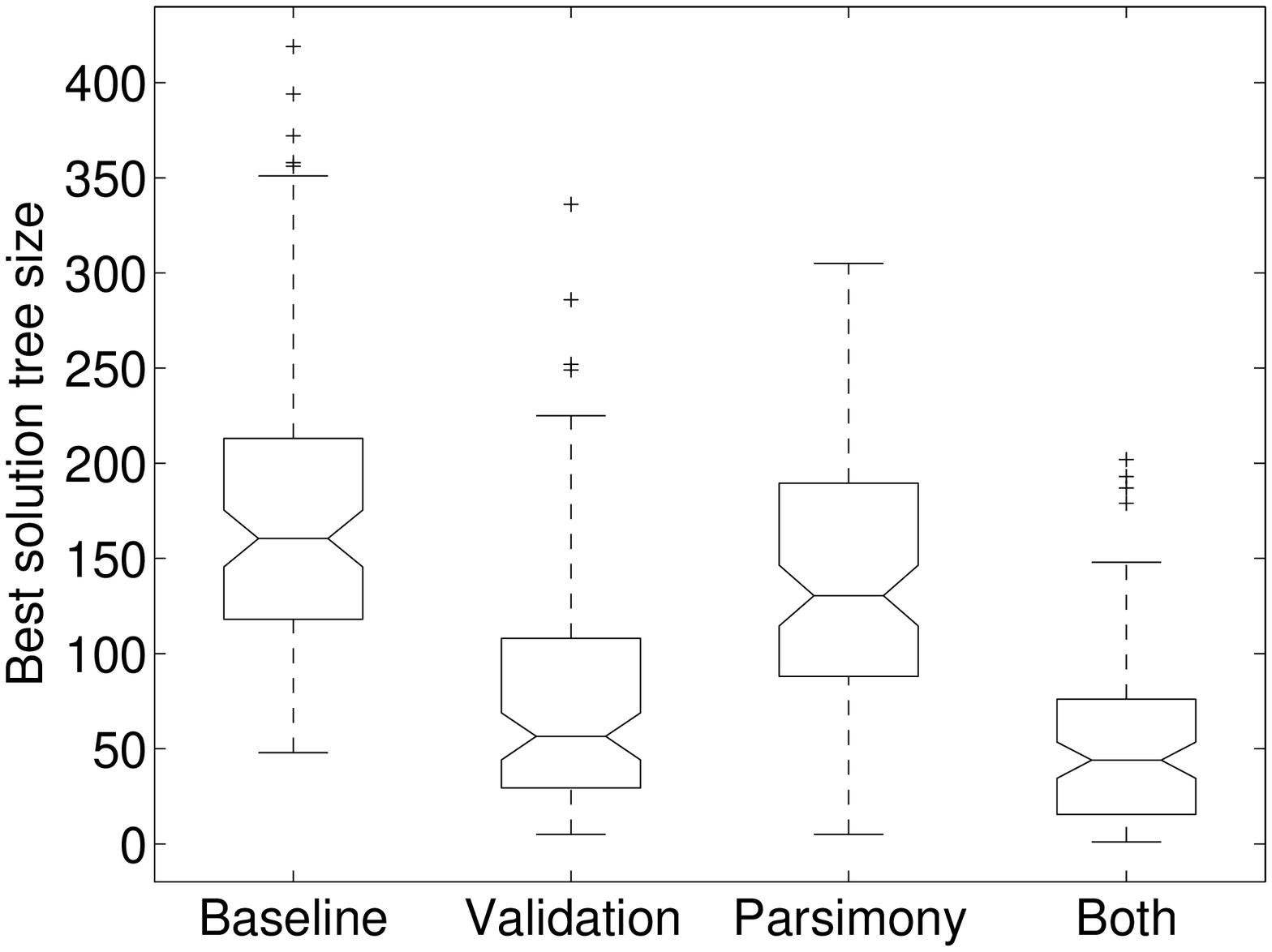}\\
bcw & cmc & ger\\
\includegraphics[width=0.33\linewidth]{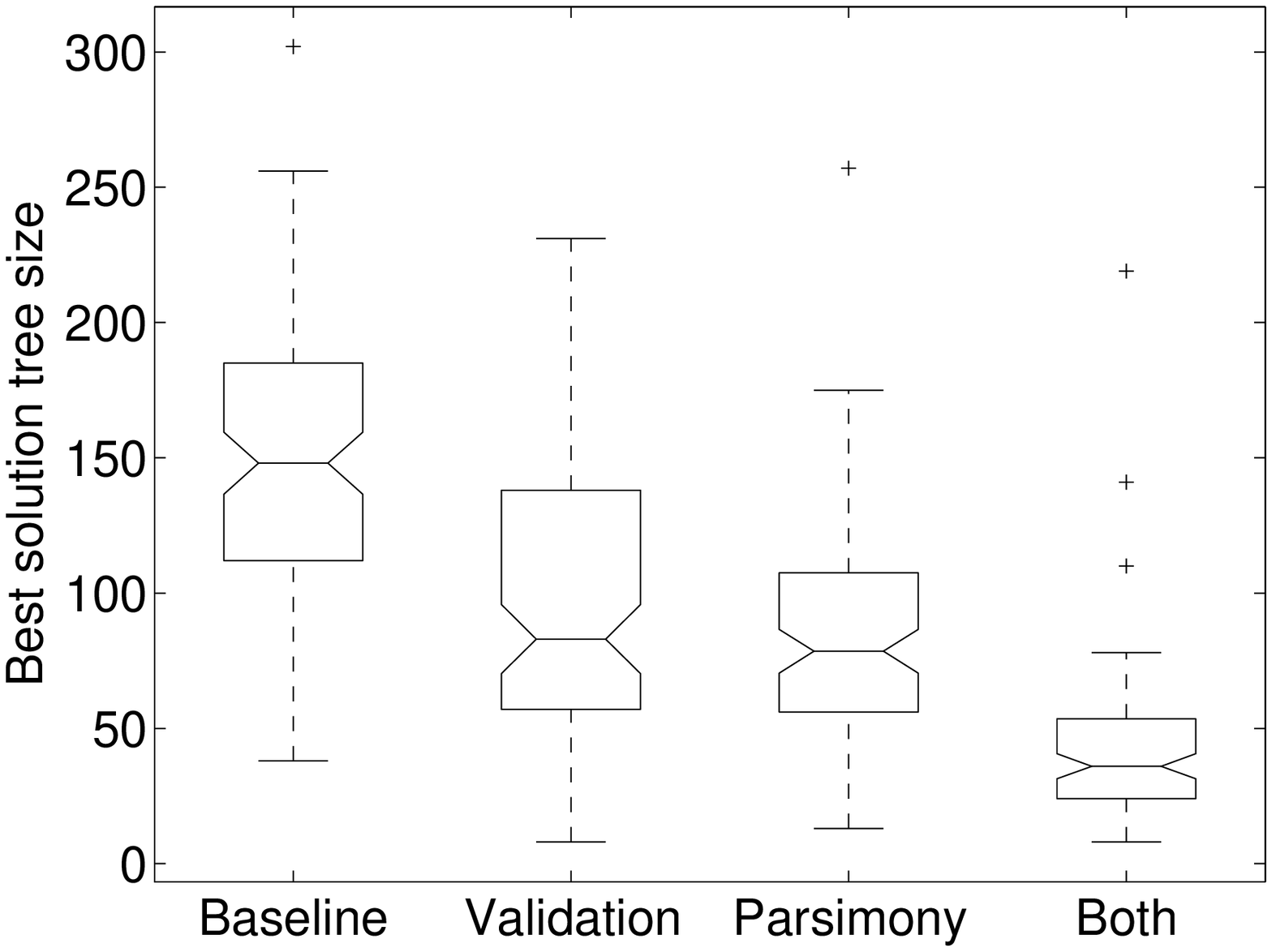} &
\includegraphics[width=0.33\linewidth]{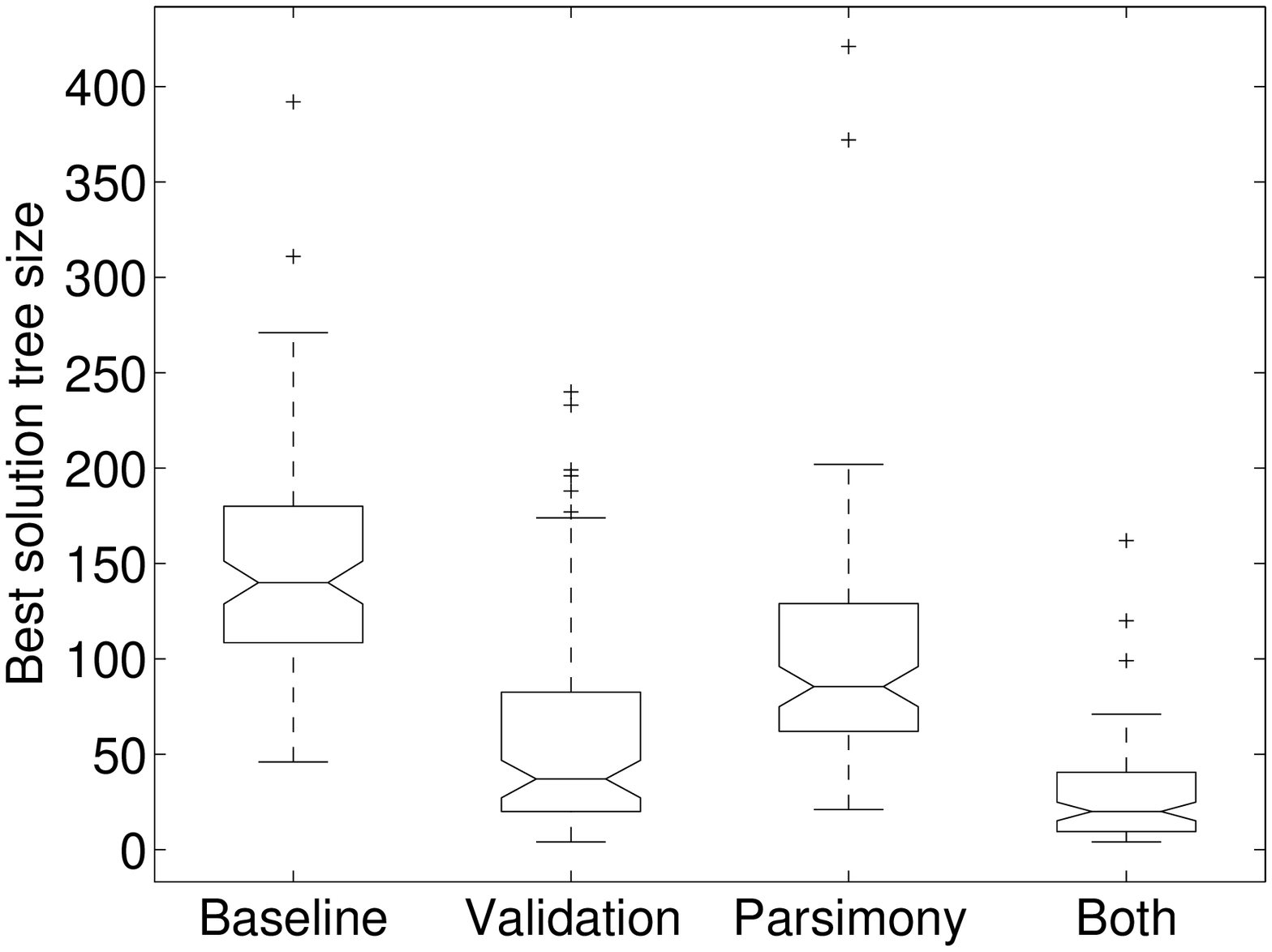} &
\includegraphics[width=0.33\linewidth]{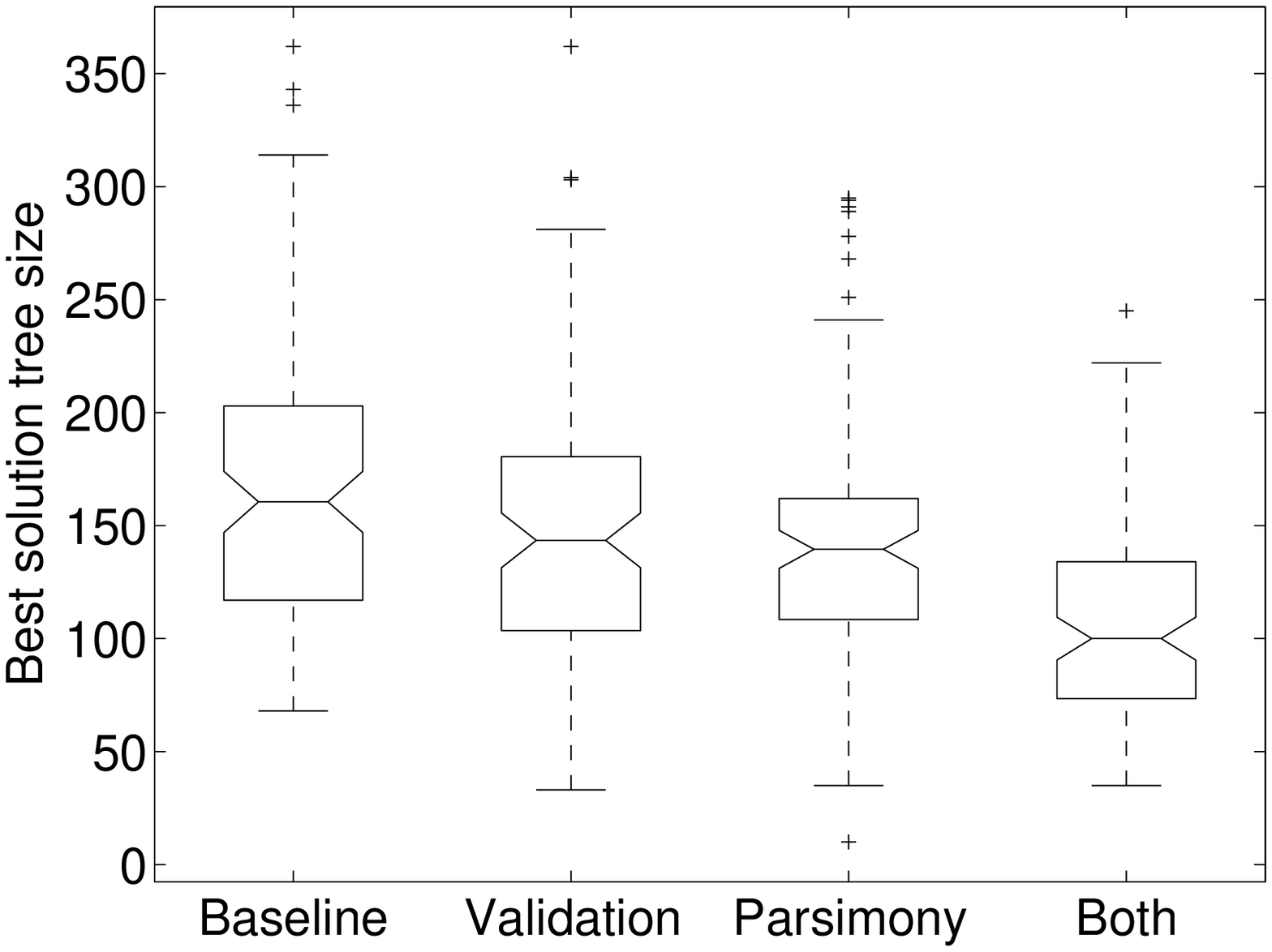}\\
ion & pid & spa\\
\end{tabular}
\end{center}
\end{figure}

Figure \ref{fig:TreeSizes} presents the one-way ANOVA box plots for the best-of-run tree sizes.
This time, it seems clear that the tested methods significantly reduce the best-of-run individual tree sizes for all tested data sets. It is interesting to note that the configurations with a validation set have generated significantly smaller best-of-run individual tree sizes compared with the parsimony pressure only approach. This is expected given that the validation set is directly used in the best-of-run individual selection process, while the parsimony pressure is used only to limit the tree sizes during the runs. Also, the important size reduction of the best-of-run solutions, especially noticeable with the combination of validation and parsimony pressure, is valuable when simplicity or comprehensibility is necessary for the application at hand. Finally, taking a look at the mean effort in Table \ref{tab:DetailedResults}, the reduction goes up to $50\:\%$ with the validation and parsimony pressure approach, compared to the baseline effort.

\section{Conclusion}
\label{sec:Conclusion}

In this paper, methodologies were investigated to improve GP as a learning algorithm. More specifically, using the GP-based setup for binary classification, the use of a validation set for selecting best-of-run individuals was tested, in order to pick solutions that generalize well. The effect of a lexicographic parsimony pressure was also tested, in order to avoid unnecessary complexity in the evolved solutions. Experimental results indicate that the use of a validation set improves a little the stability of the best-of-run solutions on the test sets, by maintaining accuracy while slightly reducing variance in most cases. This is important given the stochastic nature of GP, which can introduce important variations of the results, from one run to another. Moreover, it was shown that mild parsimony pressure applied during evolutions can sustain performance in general, while effectively reducing both solution size and effort. The combination of these two approaches apparently gives the best of both worlds, by reducing the variance of test set errors, simplifying drastically the complexity best-of-run solutions, and cutting down effort by half. 

As future work, still using a GP-based learning setup, it is planned to develop new stopping criteria based on the difference between training and validation set error rates. It is also planned to study the effect of changing the test cases during the course of the evolution for GP-based learning, using methods such as competitive co-evolution and boosting.

\subsection*{Acknowledgments}

This work was supported by postdoctoral fellowships from the ERCIM (Europe) and the FQRNT (Qu\'ebec) to C. Gagn\'e.

\bibliographystyle{splncs}
\bibliography{gagne-paper}

\end{document}